\documentclass[runningheads]{llncs}
\usepackage[T1]{fontenc}
\usepackage{graphicx}
\usepackage{subcaption}
\usepackage{booktabs}
\usepackage{duckuments}
\usepackage{mathrsfs}
\usepackage{enumitem}
\usepackage{cite}
\usepackage{float}
\usepackage{orcidlink}
\usepackage{makecell}
\usepackage{dblfloatfix}
\usepackage{eso-pic}

\usepackage{color}

\begin{document}
\title{Probing an Embodied LLM: When Higher Observation Fidelity Hurts Problem Solving}
\titlerunning{Probing an Embodied LLM}
\author{Oussama Zenkri\inst{1,2,3}\orcidlink{0009-0006-4169-1460} \and
Oliver Brock\inst{1,2,3}\orcidlink{0000-0002-3719-7754}}
\authorrunning{Zenkri O. \and Brock O.}
\institute{
Robotics and Biology Laboratory, Technische Universität Berlin, Germany \and
Science of Intelligence, Research Cluster of Excellence, Berlin, Germany\thanks{Funded by the Deutsche Forschungsgemeinschaft (DFG, German Research Foundation) under Germany's Excellence Strategy - EXC 2002/1 ``Science of Intelligence'' - project number 390523135.} \and
Robotics Institute Germany (RIG)}
\maketitle     
\AddToShipoutPictureFG*{%
  \AtPageLowerLeft{%
    \raisebox{10mm}{%
      \hspace*{\dimexpr1in+\hoffset+\oddsidemargin}%
      \parbox{\textwidth}{%
        \footnotesize
        \textcopyright{} Springer Nature Switzerland AG 2026\par
        Accepted at \textit{From Animals to Animats: 18th International
        Conference on the Simulation of Adaptive Behavior (SAB) 2026}.
        DOI:\href{https://doi.org/10.1007/978-3-032-39953-3_25}
{10.1007/978-3-032-39953-3\_25}
      }%
    }%
  }%
}       
\begin{abstract}

Large Language Models (LLMs) are increasingly proposed as cognitive components for robotic systems, yet their opaque decision processes make it difficult to explain success or failure in closed-loop embodied tasks. Following an empirical AI methodology, we study an embodied LLM agent behaviorally by varying the available information and measuring the resulting changes in behavior. Using the Lockbox, a sequential mechanical puzzle with hidden interdependencies, we evaluate LLMs across RGB, RGB-D, and ground-truth symbolic observations in a physical robotic setup and use simulation to probe the resulting behavior. Counterintuitively, agents perform best under raw RGB input and worst under perfect ground-truth observations. In simulation, we probe this effect by randomly flipping perceived action outcomes and find that moderate noise improves performance, peaking at a 40\% flip probability with a 2.85-fold success rate increase over the noise-free baseline. Further analysis links this gain to a reduction in repetitive action loops. These findings suggest that success rates alone are insufficient for evaluating LLMs, as measured performance may reflect the interaction between perceptual errors and reasoning failures rather than robust problem solving.
\vspace{-0.5em}
\end{abstract}

\let\thefootnote\relax\footnotetext{ChatGPT 5.5 and Claude Sonnet 4.6 were used to assist with text improvement. The lockbox and robotic arm sketches in Fig.~\ref{fig:experimental_setup} were generated using ChatGPT 5.5.}

\section{Introduction}

Large Language Models are increasingly proposed as cognitive components for robots operating in unstructured environments, where task execution requires more than perception or language understanding alone~\cite{mon2025embodied, ahn2022can}. Many robotic tasks demand sequential problem solving over extended interactions: agents must integrate observations over time, infer hidden structure, and adapt their behavior based on action outcomes. These abilities are key to embodied intelligence, where problem solving unfolds as a closed-loop process between agent and environment.

Analyzing embodied LLM agents requires moving beyond outcome-based evaluation toward methods that reveal the behavioral mechanisms underlying success and failure. However, LLMs are largely opaque, as their reasoning processes remain inaccessible, and the providers of the most capable models do not expose their weights. In embodied settings, performance further reflects the coupled effects of perception, reasoning, and environmental dynamics. A success or failure therefore cannot be attributed directly to reasoning ability alone, and aggregate success rates provide only a limited view of performance. They do not reveal whether an agent solved a task through coherent inference, accidental exploration, perceptual error, or an interaction between these effects.

We adopt a behavioral probing approach, aligned with empirical AI methodology~\cite{cohen1995empirical} and inspired by the logic of system identification: rather than estimating an explicit dynamical model, we perturb an opaque agent and measure how its behavior changes. We use observation fidelity to denote how faithfully an agent's observations represent the true state of its environment. Observation fidelity is a useful intervention variable because it affects the information on which the model bases its decisions while leaving the underlying task structure unchanged.

We study this problem using the Lockbox, a sequential mechanical puzzle widely used to study problem solving in biological agents and increasingly useful as a diagnostic task for embodied AI systems. The task captures key challenges of real-world problems: actions have state-dependent outcomes, and optimal actions depend on the integration of past interactions. We present the same task through different observation channels, ranging from raw RGB and RGB-D inputs to symbolic ground-truth state descriptions, and evaluate LLM agents in a physical robotic setup and use controlled simulation to probe the resulting behavior.

Our results reveal a counterintuitive pattern: increasing observation fidelity does not improve performance. Agents perform best with raw RGB input and worst when given symbolic ground-truth state. To probe this effect, we introduce controlled observational noise in simulation by randomly flipping perceived action outcomes. We find that moderate noise improves success relative to the noise-free baseline, whereas higher noise levels degrade performance again. Further analysis suggests that this improvement is associated with a reduction in repetitive action loops, a failure mode in which the agent repeats near-identical action subsequences despite making little or no progress~\cite{lai2026specra}.

Taken together, these findings show that observations are not merely passive inputs to an embodied reasoning process. They actively shape the dynamics of decision making. In closed-loop interaction, measured performance may therefore conflate perceptual errors, reasoning failures, and environmental dynamics rather than reflect robust problem solving. This highlights the need to evaluate embodied LLM agents not only by final success rates, but by probing the behavioral mechanisms that give rise to success or failure.

\section{Methodology}

To study embodied LLM agents as opaque systems, we need a task that is both physically grounded and experimentally controllable. The task must expose the agent to key challenges of real-world problem solving, while also allowing the observation channel to be varied without changing the underlying reasoning problem. This section first introduces the Lockbox as such a diagnostic task, then describes the physical and simulated environments used to study agent behavior under different observation conditions.

\begin{figure}[htbp]
  \centering
  \includegraphics[width=\textwidth]{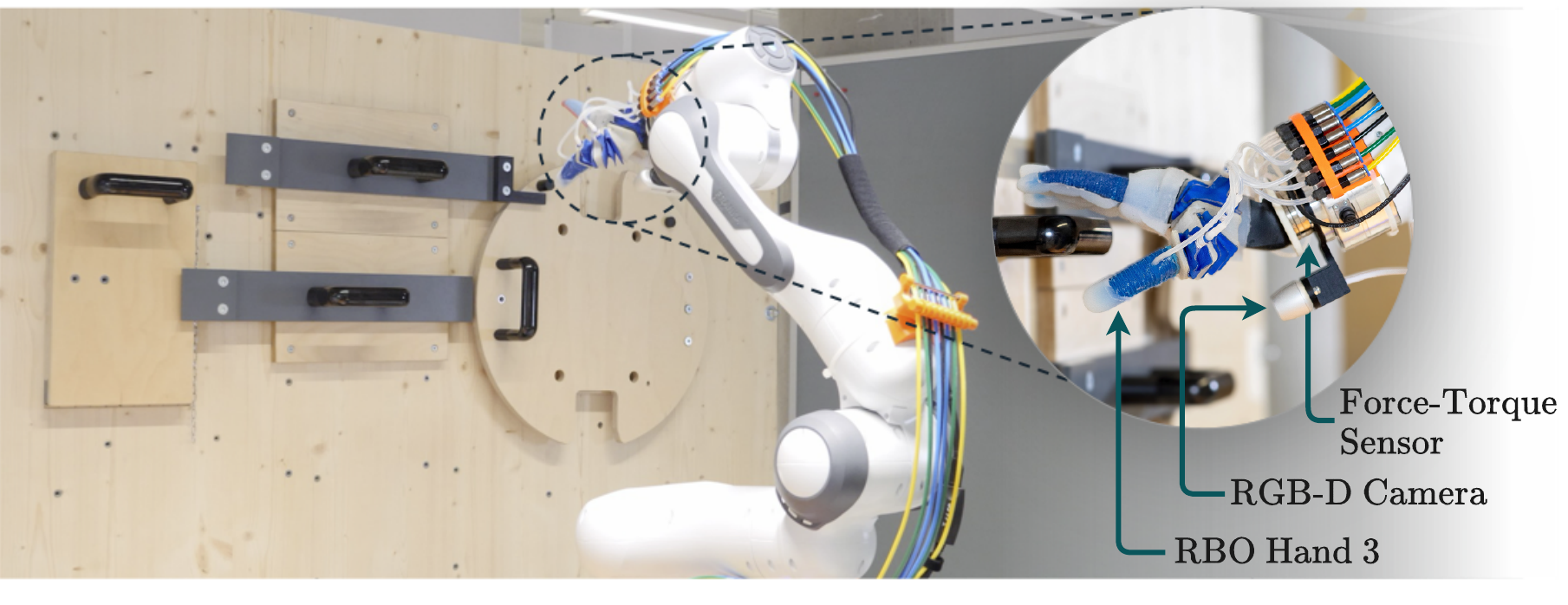}
  \caption{Our robotic system manipulating the Lockbox. Our Lockbox comprises two prismatic joints (sliding bars in the middle) and two revolute joints. The Lockbox is unlocked when the leftmost revolute joint, which we refer to as the target joint, is pulled. The robot employs a soft end-effector for manipulating the joints, an RGB-D camera for acquiring visual data, and a force-torque sensor for assessing the joint movability and guiding their manipulation.}
  \label{fig:title_figure}
  \vspace{-2em}
\end{figure}

\subsection{The Lockbox}

Lockboxes are mechanical puzzles where the objective is to gain access to a reward by manipulating a set of moving parts (called joints) that are constrained in a structured but non-obvious way. Solving them requires discovering the underlying constraint pattern and executing a sequence of actions accordingly. 

Lockboxes have been widely used as an experimental paradigm to study problem-solving behavior across a range of domains, including cognitive biology~\cite{auersperg2013explorative, lang2023challenges, stanton2024wild}, cognitive psychology~\cite{zenkri2025humanexp}, and robotics~\cite{baum2017opening, verghese2023using, liu23busybot}. The sequential, constraint-driven nature of lockboxes makes them a natural testbed for probing the behavior of LLM agents in embodied problem-solving settings. Moreover, the same underlying task can be presented across observation modalities (symbolic state descriptions, RGB, or RGB-D) without altering the latent reasoning challenge, making lockboxes a useful diagnostic task for probing how changes in the observation channel alter closed-loop problem-solving behavior.

\vspace{-0.5em}
\subsubsection{Physical Lockbox}

Our physical setup, illustrated in Fig.~\ref{fig:title_figure}, consists of four joints mounted on a wooden panel. All joints have binary states, meaning they can only occupy one of two possible positions: either endpoint of their motion range. The movability of each joint depends on the state of the others, as dictated by the underlying interlocking mechanism.

Crucially, this dependency structure is hidden from the agent and must be inferred from the relationship between actions and outcomes. The physical setup therefore provides a setting in which observation, action execution, and subsequent model decisions are tightly coupled. After each selected action, the robot manipulates the corresponding joint and returns a new observation, allowing us to study how the agent updates its behavior from interaction history.

The physical Lockbox also exposes the agent to sources of uncertainty that are difficult to reproduce in simulation, including sensor noise, contact variability, and imperfect manipulation. These factors can affect the agent's ability to infer latent dependencies, making the task qualitatively harder. The physical setup serves as the real-world setting in which behavioral phenomena can be observed before specific factors are isolated through controlled interventions in simulation.

\subsubsection{Lockbox Simulation}

Besides the physical Lockbox setup, we employ a simulated Lockbox as a controlled intervention environment: it isolates the underlying reasoning problem from the perceptual and actuation variability of the physical setup, while allowing the agent's observation channel to be perturbed systematically. The simulation preserves the essential properties of the physical setup, including the number of joints and the interlocking structure. The LLM acts at the symbolic level, receiving a description of the joints' state and selecting actions accordingly. Since the simulation provides no visual cues about the dependency structure, this structure can only be inferred from past interactions, making it suitable for testing hypotheses generated from the physical experiments.

\subsection{Experimental Setup}
\label{sec:exp_setup}

\newcommand{\thefootnote}{\fnsymbol{footnote}}

The experiments are conducted using an existing robotic manipulation system~\cite{li2024biologically}, consisting of a Franka Emika Panda robotic arm equipped with an RBO Hand 3 soft end-effector~\cite{puhlmann2022rbo}, an RGB-D camera for visual observation, and a force-torque sensor for manipulation feedback (see Fig.~\ref{fig:title_figure}).

\begin{figure}[ht]
  \centering
  \includegraphics[width=\textwidth]{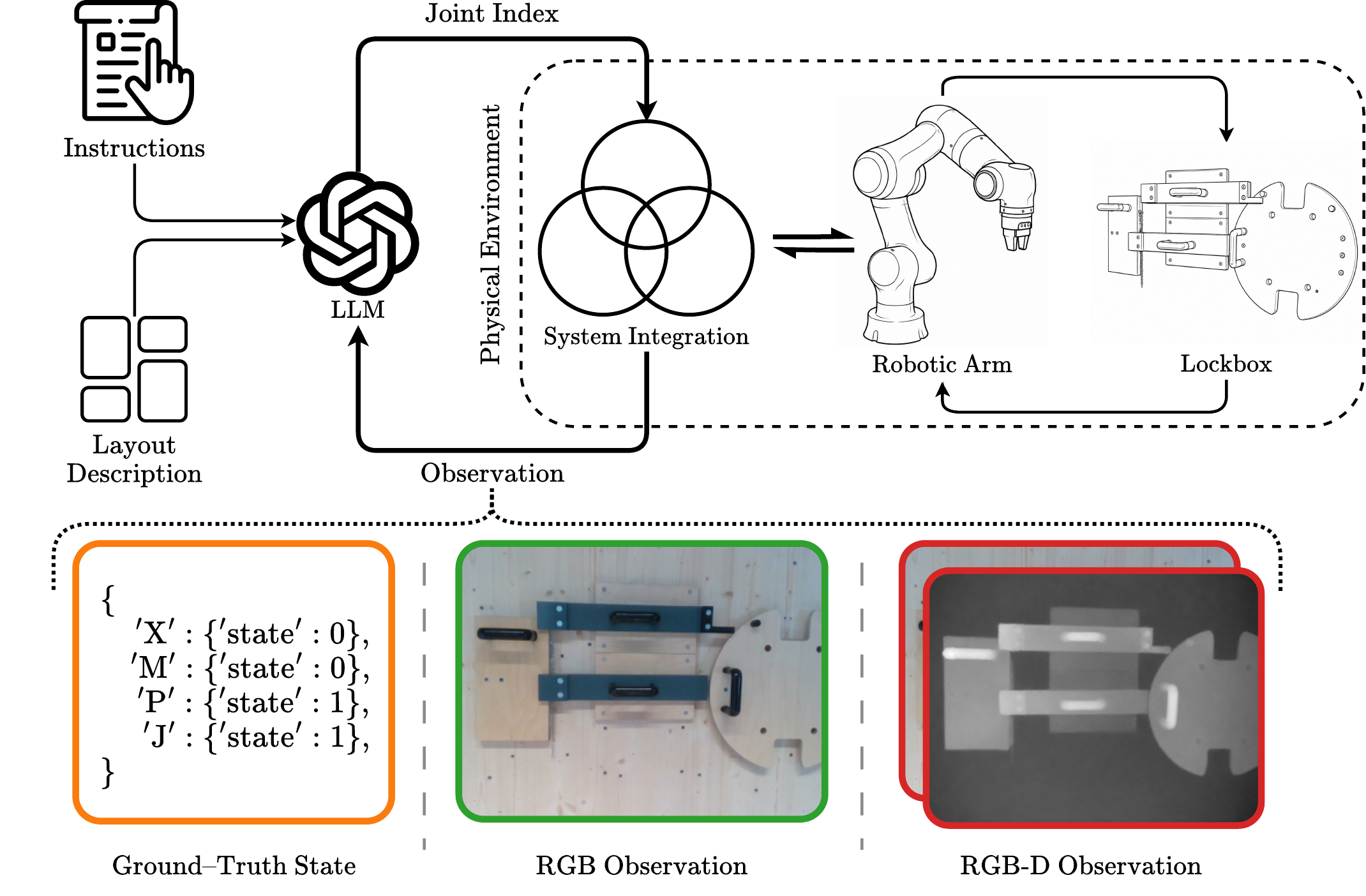}
  \caption{Experimental setup with three input modalities for the LLM. Through our system integration, the LLM interacts with the Lockbox by selecting a joint to be manipulated, after which the corresponding new observation is forwarded to the model. Bottom: three instances of the considered input modalities. The ground-truth state (left) consists of a structured representation specifying the predefined labels of each Lockbox joint and its corresponding state. The RGB observation (middle) is captured before each interaction from a fixed viewpoint, enabling direct comparison of successive observations to infer action outcomes. The RGB-D observation (right) augments the RGB modality with an aligned depth channel, providing additional geometric information about the scene.}
  \label{fig:experimental_setup}
  \vspace{-1em}
\end{figure}

A system integration layer mediates all interactions between the LLM and the physical environment, translating the model's joint selections into physical manipulations, as shown in Fig.~\ref{fig:experimental_setup}. Before each action, the robot moves to a homing position, from which a new observation of the Lockbox is captured and forwarded to the LLM, ensuring consistent viewpoints across all steps.

We employ OpenAI o1 as the decision-making LLM, selected for its reasoning capabilities in long-horizon tasks. We evaluate the model under three input modalities with varying levels of observation fidelity: (i) RGB images, (ii) RGB-D images, which add depth to the visual signal, and (iii) the ground-truth symbolic state of all joints, which bypasses the vision pipeline entirely. We treat these modalities as interventions on the agent's observation channel: the underlying task and action space remain fixed, while the information available for decision making changes. In the visual modalities, the model must infer action outcomes by comparing successive observations. In the ground-truth condition, the task reduces to symbolic reasoning over explicitly provided state descriptions.

Each trial begins with an instruction prompt describing the Lockbox problem, the task objective, and the Lockbox layout (i.e., relative joint positions and their handles' orientations).\footnote{Supplementary material, including the prompts used in this study, is available at \url{https://rbo.gitlab-pages.tu-berlin.de/papers/zenkri-sab-26/}} The instructions and layout description are identical across trials; the only modality-specific adjustments concern the observation format. All trials start from the same initial Lockbox configuration. A trial proceeds as a sequence of observation-action steps, with a maximum budget of 20 steps. Failure to move the target joint within this limit is counted as an unsuccessful trial. The agent returns a structured response containing a brief description of its observation, the selected joint label, and a concise rationale behind its decision. Since o1 fixes temperature and top-p at 1 and does not expose these parameters, the model's outputs are inherently stochastic. We therefore conduct 10 independent trials per modality to obtain reliable performance estimates.

\section{Results and Discussion}

We now use the Lockbox to identify how changes in observation fidelity affect the behavior of embodied LLM agents. The goal is not only to compare final task success across input modalities, but to understand which behavioral mechanisms may explain the observed differences. We first analyze performance in the physical robotic setup, then test the role of observational noise in simulation, and finally examine repetitive action loops as a candidate behavioral mechanism underlying the resulting performance trends.

\subsection{Observation Fidelity Alters Closed-Loop Behavior}

We first use input modality as an intervention on the agent's observation channel (Section~\ref{sec:exp_setup}), asking whether more faithful observations lead to more effective problem solving, and whether deviations from this expectation reveal systematic behavioral weaknesses. For reference, we include a human-inspired strategy for solving the Lockbox~\cite{li2024biologically}, which receives the ground-truth symbolic state and infers the dependency structure through trial and error. The heuristic is derived from prior behavioral data on human problem solving in Lockbox-like tasks~\cite{zenkri2024extracting}.

\begin{figure}[ht]
  \centering
  \includegraphics[width=\textwidth]{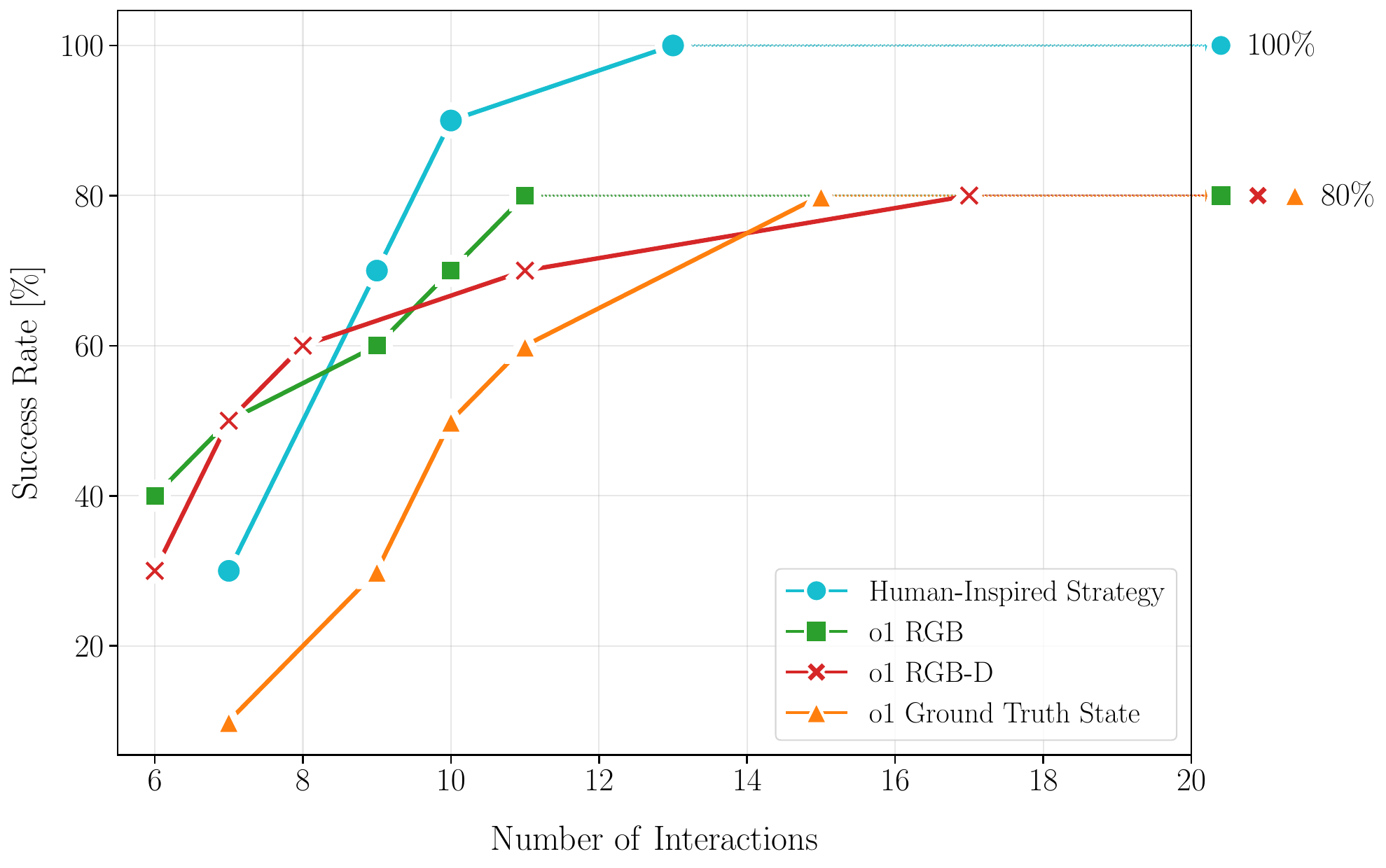}
  \caption{The human-inspired strategy outperforms o1 across all input modalities. The figure shows the success rate as a function of the number of interactions, comparing the human-inspired strategy to o1 under different input modalities. Higher input fidelity is associated with lower performance. The model performs worst under ground-truth state observations, suggesting limitations in its ability to use structured state information for decision making.}
  \label{fig:real_world_results}
  \vspace{-1em}
\end{figure}

As shown in Fig.~\ref{fig:real_world_results}, which plots cumulative success rate as a function of interaction steps, the human-inspired strategy achieves a 100\% success rate and requires fewer steps than o1 under any input modality. By contrast, o1 reaches a maximum success rate of 80\% across all three modalities. However, identical final success rates mask substantial differences in efficiency: under RGB input, o1 reaches 80\% success after 11 steps, whereas the ground-truth state input requires 15 steps. The human-inspired strategy reaches the same success rate within 9 steps. These differences show that final success rate alone provides an incomplete picture of agent performance because it does not capture the efficiency or structure of the decision-making process that produced the outcome.

Turning to the effect of input modality, the results run counter to the common assumption that more accurate observations should improve performance. o1 performs best with RGB input and worst with ground-truth state, with RGB-D falling in between. In Fig.~\ref{fig:real_world_results}, the RGB condition reaches most success thresholds faster than RGB-D, which in turn mostly outpaces the ground-truth condition at intermediate thresholds. These results suggest that additional or more accurate input is not only underutilized but may hinder problem solving. One explanation for this observation is that the image input contains more information than the ground-truth state description, which the model could use to better infer the interdependencies between the joints, as argued in prior work on visually grounded LLM planning~\cite{hu2023look}. However, the misinterpretation rates in Table~\ref{tab:misinterpretations} and the controlled experiments in Section~\ref{sec:noise_injection} point toward a different explanation.

To better understand this effect, we analyze the model's ability to correctly identify state changes resulting from its actions. We define a misinterpretation as an incorrect assessment of the state transition following a manipulation. 
For visual inputs, misinterpretation rates vary substantially across trials and modalities, with the large variability suggesting inconsistent perceptual grounding rather than a systematic failure mode (see Table~\ref{tab:misinterpretations}). Why depth information degrades performance relative to RGB remains unclear. The additional depth channel may shift the model's attention toward task-irrelevant spatial features.

\begin{table}[bp]
\vspace{-2em}
\setlength\tabcolsep{1.1em}
\begin{center}
\caption{State Misinterpretation Rates Across Input Modalities}\label{tab1}
\begin{tabular}{lll}
\toprule
\textbf{Input Modality} & \textbf{\makecell[l]{Average State Misinterpreta-\\tion Rate per Trial}} & \textbf{\makecell[l]{Total State\\Misinterpretations}}\\
\midrule
\textbf{RGB} & $21.7\%\footnotesize\pm12.2\%$ & 22$^{\ast\ast}$ \\
\textbf{RGB-D} & $26.8\%\footnotesize\pm12.5\%$ & 35$^{\ast\ast}$ \\
\textbf{Ground-Truth} & $4.9\%\footnotesize\pm6.9\%$ & 6$^{\ast}$ \\
\bottomrule
\label{tab:misinterpretations}
\end{tabular}
\end{center}
\vspace{-1.8em}
~~~$\ast:$ \footnotesize occurred in 4 out of 10 trials
~~~$\ast\ast:$ \footnotesize occurred in all 10 trials
\vspace{-1em}
\end{table}

Surprisingly, misinterpretations also occur in the ground-truth condition, yielding an average step-level error rate of approximately 5\%. Since the ground-truth input requires no perceptual inference, these errors instead indicate failures in reasoning over structured representations, most plausibly through hallucinating state changes, which contradicts the provided input. 

These state misinterpretations suggest that the advantage of the visual modalities over ground-truth state may not stem from additional scene information, but rather from naturally occurring perceptual noise that proves beneficial to the model's decision making. If lower-fidelity visual inputs improve performance by introducing stochasticity into the perceived state, then a similar effect should appear when state uncertainty is introduced directly in a controlled symbolic simulation. The following section tests this prediction.

\subsection{Perturbing Observations Reveals a Non-Monotonic Response}
\label{sec:noise_injection}

To test the hypothesis above, we design a controlled experiment that removes visual observations entirely. We use a simulated Lockbox with a ground-truth state description as model input and introduce noise directly into the observations through random joint-state flips applied at controlled probabilities. Specifically, with a probability $p$, the agent receives the inverted outcome of the action it took. 
This perturbation affects only what the agent perceives, not the actual Lockbox state, thereby mimicking the effect of misinterpreting a visual observation.

\begin{figure}[ht]
  \centering
  \includegraphics[width=\textwidth]{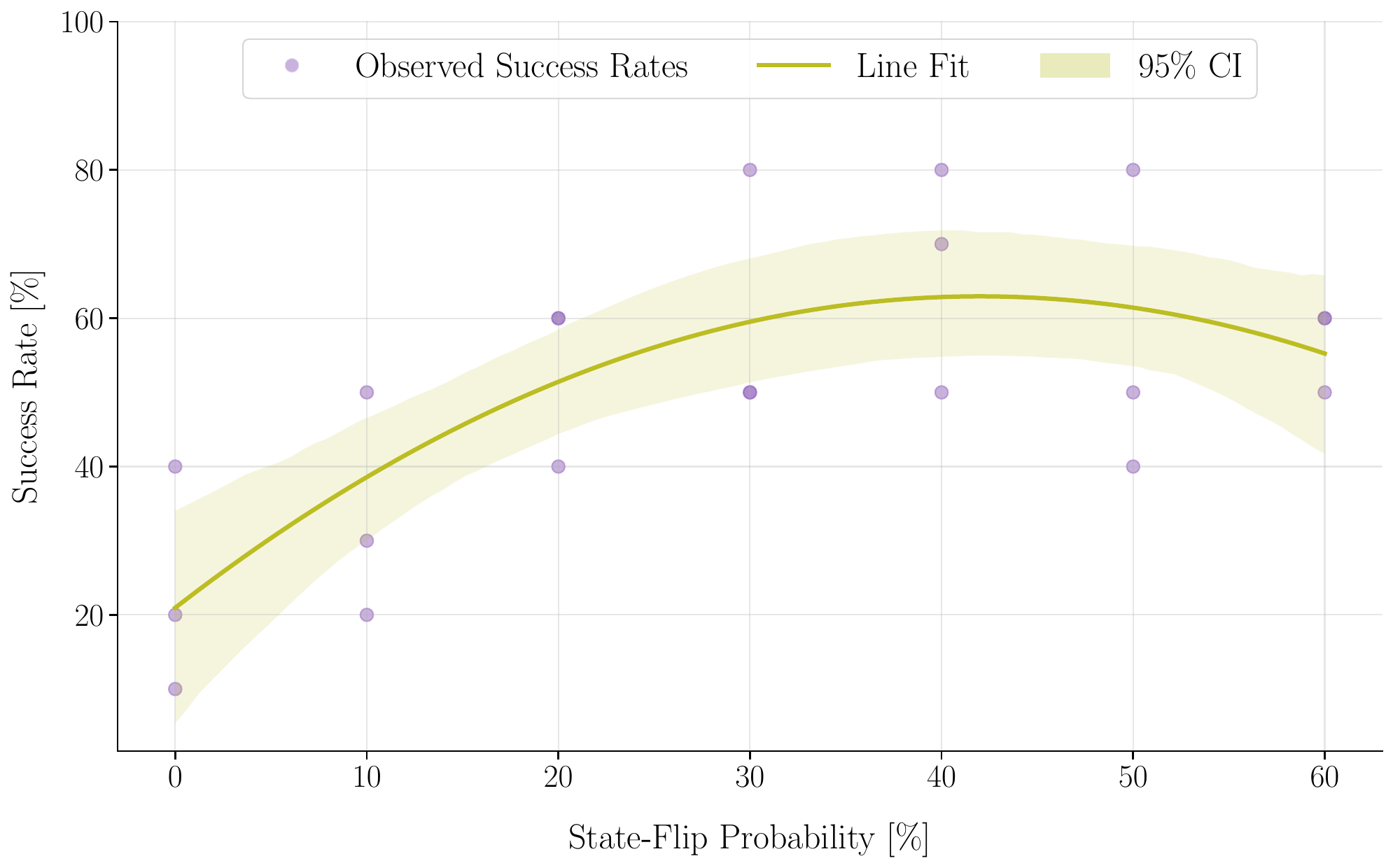}
  \caption{Performance exhibits a non-monotonic dependence on perceptual noise. Success rates increase with state uncertainty, peaking at a 40\% state-flip probability ($\approx2.85\times$ the ground-truth baseline, on average), before declining at higher noise levels, suggesting that moderate noise can improve the observed task performance without necessarily reflecting improved reasoning.}
  \label{fig:random_state_flip}
  \vspace{-1em}
\end{figure}

We evaluate state-flip probabilities ranging from 0\% to 60\% in increments of 10\%, examining how varying levels of observational noise affect the model's ability to solve the Lockbox. For each flip probability, the experiment is repeated three times, with each repetition consisting of 10 independent trials (210 trials in total). We use GPT-4o for this study because the experiment operates purely over symbolic states rather than multimodal observations, and because the larger number of controlled intervention trials would otherwise be prohibitively costly. Accordingly, the simulation results should be interpreted as evidence for a plausible mechanism, not as a direct explanation of the physical results.

Figure~\ref{fig:random_state_flip} shows the observed success rates across all repetitions for each state flip probability, together with a second-order polynomial fit to better capture the underlying trend (the order of the fit was determined through AIC and cross-validation RMSE minimization). The fit reveals a non-monotonic relationship between observational noise and task performance: starting from an average success rate of 23.3\% under ground-truth observations, performance rises to a peak at a flip probability of 40\% before declining at higher noise levels. The improvement relative to the noise-free baseline indicates that moderate observational noise can improve measured task performance in this controlled setting.

These results support the hypothesis that the performance gap between observation modalities is in part linked to stochasticity in the perceived state, rather than to visual information alone. However, they offer no mechanistic explanation for why perceptual noise should improve performance in the first place. In the following section, we investigate a mechanism that may underlie this effect.

\subsection{Repetitive Action Loops as a Candidate Failure Mode}
\label{sec:action_loops}

A recurring behavioral pattern observed across trials is the tendency of the LLM to produce repetitive action subsequences within a single trial. For example, an agent might select joint A, then joint B, then joint C, before looping back to the same subsequence at later timepoints of the trial. Although the agent eventually breaks out of such loops, the intervening redundant actions substantially degrade the problem-solving efficiency and frequently cause the agent to exhaust its interaction budget before solving the Lockbox.

This observation raises the question of whether observational noise is associated with fewer repetitive action loops, and thereby increases the agent's probability of success. To evaluate this hypothesis, we examine whether the occurrence of action loops within a trial's action sequence is correlated with success rate, and whether this occurrence is modulated by the state-flip probability.

\begin{figure}[ht]
  \centering
  \includegraphics[width=\textwidth]{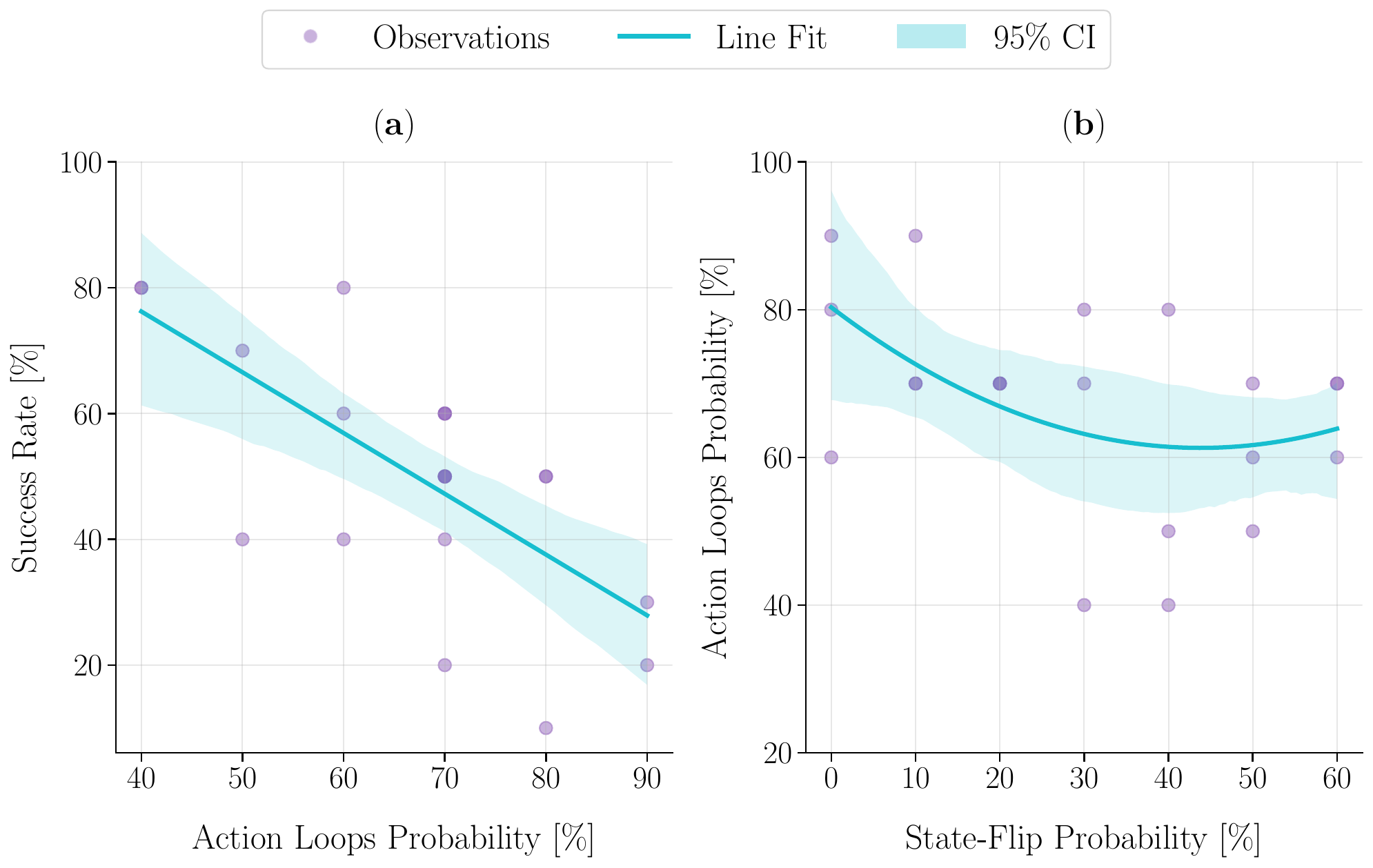}
  \caption{Perceptual noise is associated with fewer action loops, which are associated with reduced performance. (a) Action loop probability and success rate are negatively correlated ($r=-0.69$, Pearson correlation), with higher engagement in action loops corresponding to lower success rates. (b) The probability of engaging in repetitive action sequences decreases with increasing perceptual noise, reaching its minimum at a 40\% state-flip probability (on average, $\approx 26\%$ lower than the ground-truth baseline), before rising again at higher noise levels. Together, these trends suggest that the observed increase in success rates under moderate noise is consistent with the reduction in repetitive action sequences.}
  \label{fig:action_loops}
  \vspace{-1em}
\end{figure}

We define an action loop as any subsequence of actions of length three or higher that appears at least twice within the same trial. For a given trial, the set of action loops covering that trial is determined as the non-overlapping collection of individual action loops that maximizes the total coverage of the action sequence in that trial. This combinatorial optimization problem is solved via binary Integer Linear Programming (ILP), applied over the set of action loops obtained by greedily enumerating all repeated subsequences in the trial.

Figure~\ref{fig:action_loops}$(a)$ shows that action loop occurrence is negatively correlated with success rate: as the proportion of trials containing action loops increases, success rate decreases consistently. Figure~\ref{fig:action_loops}$(b)$ further shows that this occurrence is non-monotonically dependent on the state-flip probability. It decreases from a maximum average value of approximately 80\% at zero noise, reaches a minimum near a state-flip probability of 40\%, reflecting a decrease of approximately 26\%, and then rises again at higher noise levels. This pattern is consistent with the hypothesis that moderate perceptual noise reduces the likelihood of action loop formation and may contribute to the observed improvement in task success.

Becoming trapped in repetitive loops that repeatedly generate near-identical action sequences is a well-documented failure mode of LLM-based agents~\cite{lai2026specra, yao2022react}. In interactive decision-making settings like the Lockbox, the tendency is compounded when repeated failed attempts to move a joint leave the environment state unchanged, causing the agent to observe the same state across consecutive steps and providing no novel contextual signal to drive exploration toward alternative actions. This interpretation aligns with prior findings that noise injected into the prompt can increase the diversity of LLM responses~\cite{agrawal2026diversity}.

\section{Limitations}

First, our evaluation was restricted to models from a single provider, OpenAI. Given the field's rapid development, these models no longer represent the state of the art. Whether the observed behavior generalizes to newer models, to models from other providers, or to open-source models remains an open question. Cross-provider comparison was deliberately outside the scope of this study. Future work should therefore assess whether our findings reflect a broader limitation of LLM-based agents or are specific to the models evaluated here.

Second, our evaluation was conducted on a single visual instance of the Lockbox, meaning that the observed behavior could, in principle, be specific to this particular layout. However, other simulated experiments with actively varied visual layouts~\cite{zenkri2026how} provide preliminary evidence that the findings are not layout-specific. A systematic study of visual variation in real-world settings is nonetheless needed to establish how broadly these findings generalize.

Finally, model performance was benchmarked against a human-inspired strategy rather than empirical data from human participants solving the same task. Although this heuristic provides a principled approximation of human reasoning, it may not fully capture the variability, strategies, and error patterns of actual human performance. A direct comparison with human participants would therefore provide a stronger and more ecologically valid benchmark.

\section{Conclusion}

This work set out to probe the behavior of LLMs as cognitive components in embodied systems. Specifically, we examined how their performance changes across different levels of observation fidelity, since real-world perception is inherently noisy. Rather than improving monotonically with increasing fidelity, performance peaks under moderate perceptual noise and degrades when agents are given perfect observations. 
This counterintuitive finding is consistent with a candidate mechanism: accurate feedback may sustain repetitive action loops, whereas erroneous observations can disrupt them. In other words, what looks like better performance under lower-fidelity inputs should not be interpreted as evidence for stronger reasoning, but may instead reflect a side effect of compounding perceptual errors that happen to be beneficial for this task. This distinction matters beyond the Lockbox. In embodied settings where perception and reasoning are simultaneously imperfect and difficult to disentangle, success rates may reflect the accidental interplay of these imperfections as much as genuine problem-solving ability. 
We conclude that the evaluated models are still subject to weaknesses at both the perceptual and reasoning levels, and that evaluating LLMs rigorously requires real-world closed-loop benchmarks combined with controlled behavioral interventions that reveal the mechanisms behind success and failure.

%
% ---- Bibliography ----
%
\bibliographystyle{splncs04}
\bibliography{bibliography}

\end{document}